\documentclass[runningheads]{llncs}
\usepackage{graphicx}
\usepackage{url}
\usepackage{multirow}
\usepackage[normalem]{ulem}
\useunder{\uline}{\ul}{}
\usepackage{subfigure}
\usepackage{algorithm}
\usepackage{algpseudocode}
\usepackage{amsmath,bm}

\begin{document}
	%
	\title{MRNet: a Multi-scale Residual  Network  for  EEG-based Sleep Staging}
	%
	%
	\author{Xue Jiang\inst{1}}
	%
	%
	\institute{Wuhan University, Wuhan, China\\
	\email{jxt@whu.edu.cn}}
	%
	\maketitle              
	\begin{abstract}
		Sleep staging based on electroencephalogram (EEG) plays an important role in the clinical diagnosis and treatment of sleep disorders. In order to emancipate human experts from heavy labeling work, deep neural networks have been employed to formulate automated sleep staging systems recently. However, EEG signals lose considerable detailed information in network propagation, which affects the representation of deep features. To 
		address this problem, we propose a new framework, called MRNet, for data-driven sleep staging by integrating a multi-scale feature fusion model and a Markov-based sequential correction algorithm. 
		The backbone of MRNet is a residual block-based network, which performs as a feature extractor.
		Then the fusion model constructs a feature pyramid by concatenating the outputs from the different depths of the backbone, which can help the network better comprehend the signals in different scales. 
		The  Markov-based sequential correction algorithm is designed to reduce the output jitters generated by the classifier. The algorithm depends on a prior stage distribution associated with the sleep stage transition rule and the Markov chain. 
		Experiment results demonstrate the competitive performance of our proposed approach on both accuracy and F1 score (e.g., 85.14\% Acc and 78.91\% F1 score on Sleep-EDFx, and 87.59\% Acc and 79.62\% F1 score on Sleep-EDF). 	

	\end{abstract}
	\section{Introduction}
	Sleep is a basic human need and is critical to both physical and mental health.
	Not getting enough sleep can lead to sleep disorders that interfere with human's normal physical, mental, social and emotional functions    \cite{Liu440}.
	Medical experts regard sleep staging as a supplementary tool for the  analysis of sleep disorders \cite{Sateia2014}.
	Sleep staging, also named as sleep score, generally follows the standard  set by the American Academy of Sleep Medicine (AASM) in 2007 \cite{Malhotra2014a}, which divides the sleep into five distinct stages: Wake, Rapid Eye Movement (REM), N1, N2 and N3 (non-REM).
	Manual sleep staging is a lengthy and tedious task that requires the continuous attention of well-trained human experts. For example, an expert needs approximately 5 hours to score 24h electroencephalogram (EEG) \cite{Malhotra2014a}.
	Therefore, it is necessary to develop a data-driven tool to realize automatic sleep staging.
	

	Some learning algorithms  have been proposed  for the automatic sleep staging  based on  EEG signals of brain activities.
	Typical methods  can be divided into two parts: handcrafted feature-based models and deep learning-based models. Handcrafted feature-based models usually extract features from EEG signals by some traditional strategies (e.g., short-time Fourier transform (STFT)   and discrete wavelet transform (DWT)   \cite{Yildiz2009}), and then feed these features into classification machines (e.g.,  support vector machine (SVM)  \cite{Alickovic2018}).
	In deep learning-based models, EEG feature extraction is automatically realized by deep neural networks (DNNs) \cite{Supratak2017}, enabling end-to-end automated sleep stage classification. For example,  \cite{Perslev2019} proposed a temporal fully convolutional network with the encoder-decoder architecture for EEG-based signal analysis, and  \cite{Seo2020} used long short-term memory network (LSTM) to learn the temporal relationship between epoch-wise features individually extracted from each epoch by convolutional neural networks (CNNs). Different from the former approaches  requiring  additional handcrafted tuning,  the deep learning-based models enjoy a total automated process for EEG-based sleep staging.
	
	Despite the attractive characterizations of  deep models, there are still some limitations for mining EEG signals.
	Processing raw EEG signals by DNNs is a procedure of data filtering and abstraction. The layer-by-layer convolution operation filters out the signal noise, and encodes the signal into features  subsequently used  for classification.
	Usually,  only the features associated with the last convolutional layer are considered for EEG-based classification, and the information loss of the feature details caused by the previous convolutional layers is ignored. The ignored features from other network levels may be beneficial for sleep staging.
	To address this issue, we integrate  a residual connection with a multi-scale feature fusion model to better utilize feature details.
	
	On a different note, the data labeling process is not uniform among different experts because of their subjectivity, resulting in labeling noises. The previous models think that sleep stage transitions are independent, and the output jitters  caused by the labeling noises tend to happen frequently. 
	This may contradict with  sleep stages under the physiological rule, e.g., sleep cycle  \cite{wiki}. For example, when you fall asleep, the change in sleep stages can be observed as W$\rightarrow$N1$\rightarrow$N2$\rightarrow$N3. In order to smooth the output jitters for improving the result's reliability, we propose a Markov-based sequential correction algorithm based on the sleep cycle.
	
	The main contributions of this paper are summarized as  follows:
	\begin{itemize}
		\item A new residual-based architecture, called  MRNet,  is proposed for automatic sleep staging on raw single-channel EEG;
		\item A feature pyramid model is formulated  to better exploit the multi-scale deep features including the previously lost information  in the forward propagation of DNNs;
		\item We combine the physiological characteristic (sleep cycle) and the Markov chain to design a Markov-based sequential  correction algorithm, which brings significant improvement on sleep staging results through increasing the consistency of the output.
	\end{itemize}
	
	\section{Proposed MRNet}
\begin{figure}[t]
		\includegraphics[width=\textwidth]{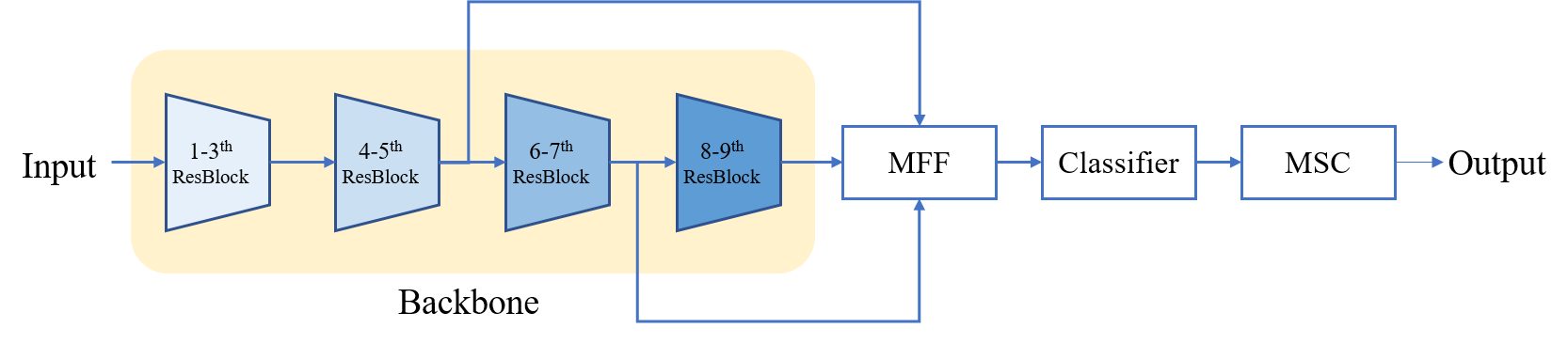}
		\caption{The overall architecture of MRNet. MFF represents the multi-scale  feature  fusion  model. MSC represents the Markov-based sequential correction algorithm.} \label{fig0}
	\end{figure}

	The overall architecture of MRNet is shown in Fig. \ref{fig0}. The residual block-based backbone extracts the features from EEG signals and sends them to the multi-scale  feature  fusion  model (MFF). Then, MFF collects and fuses the features in different scales. The fused features are used for predictions by the classifier. Finally, the Markov-based sequential correction algorithm post-processes the predicted results.
	
	\subsection{Backbone Architecture}
\begin{figure}[t]
		\includegraphics[width=\textwidth]{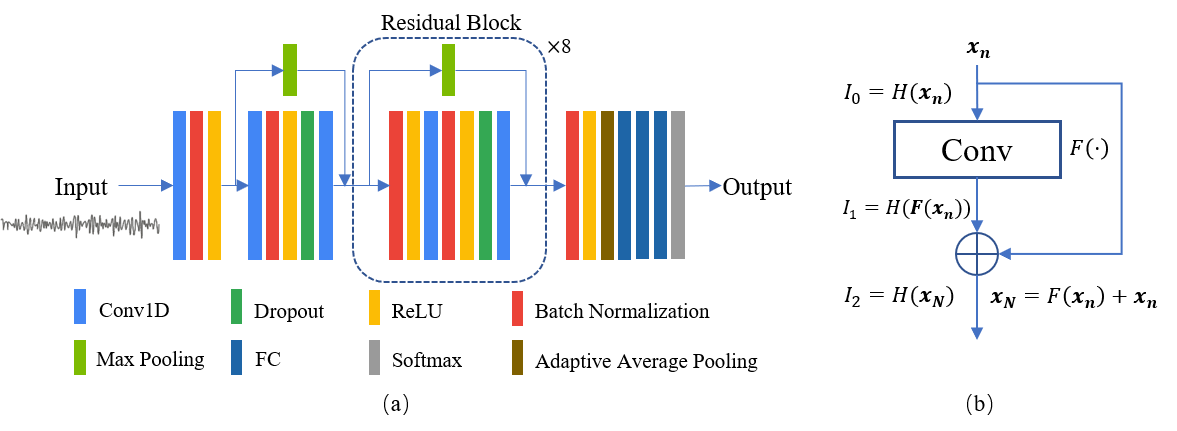}
		\caption{(a) The backbone of MRNet.   (b) Information change after a residual block.} \label{fig1}
	\end{figure}

	As shown in Fig. \ref{fig1}(a), the backbone architecture used here  is based on  a 1d residual block-based network \cite{Hannun2019}, which takes  raw EEG signals as input and outputs the estimated sleep stage.

	Let $H(\cdot)$ be the information entropy operator and let $F(\cdot)$ be the convolutional layer. The process of information change after a residual block is shown in Fig.\ref{fig1}(b).  From the viewpoint of information theory  \cite{Gray2011}, the  information metrics of input signal $\bm{x_n}$ (before the residual block) and the output (after the convolutional block) are 
	\begin{equation}
	I_0 = H(\bm{x_n})
	\end{equation}
	and
	\begin{equation}
	I_1 =H(F(\bm{x_n}))
	\end{equation}
	respectively. 
	The information of the output $\bm{x_N}$ after the residual block is
	\begin{equation}
	I_2 = H(\bm{x_N})=H(F(\bm{x_n}) + \bm{x_n}).
	\end{equation}
	According to the information processing theorem  \cite{DeGiacomo2012}, $I_0$ is the upper boundary of   $I_1$, i.e., $I_0 \geq I_1$.
	In other words, signals are likely to suffer information loss after convolutional layers.
	Therefore, the residual block adopts a skip connection to recover the upper boundary of $I_0$ to mitigate the information loss.
	
	%

	The backbone architecture of MRNet has 19 convolutional layers, consisting of 9 residual blocks  \cite{He2016} with two convolutional layers per block.
	The filter length of convolutional layers is 32 because a large convolutional layer filter brings a large reception field, which can effectively capture the temporal information of physiological signals.
	Every two residual blocks subsample its inputs by a factor of 2.
	As for the classifier, we choose Adaptive Average Pooling to subsample the length of each channel to 3.
	This effectively reduces the parameters of subsequent full connection layers and prevents overfitting by regularization.
	
	In this paper, hyperparameters of the network architecture are chosen via a combination of grid search.
	In detail, we mainly search for the number of convolutional layers, the size and number of convolutional filters, and the selection probability of dropout layers.
	
	\subsection{Multi-scale Feature Fusion}
	From the aspect of signal representation, with propagating deeper in the network, signals are further processed and reorganized.
	In this process, redundant information in EEG signals will be discarded, so that the Symbolic Entropy  \cite{Gray2011}, i.e., the information, contained within a unit length,  of EEG signals will increase.
	However, as shown in Section \textbf{2.1}, information processing and reorganization will inevitably be accompanied by information loss. We conduct signal reconstruction experiments to study this phenomenon.  Here, we select the features extracted from different depths of the backbone and use  deconvolutional layers to reconstruct signals. Experimental results reveals that  the reconstruction performance is worse when the feature goes deeper,  which verify that some information has been permanently lost in the process of the propagation.

	
	There are mainly two reasons for information loss in the network architecture. One is the feature confusion caused by convolutional layers.
	In detail, the large reception field brought by the long convolution filters will make the boundary between signals more ambiguous.
	The other is that the downsampling operation after every two residual blocks makes the time-domain information  be encoded to the channel-wise.
	Specially, the length of EEG signals decreases as it propagates forward through the network.
	In the meantime, there is also an increase in the number of feature channels, resulting in a series of features at different resolutions.
	Hence, the key strategy for reducing information loss is to fuse these features.
	
\begin{figure}[t]
		\includegraphics[width=\textwidth]{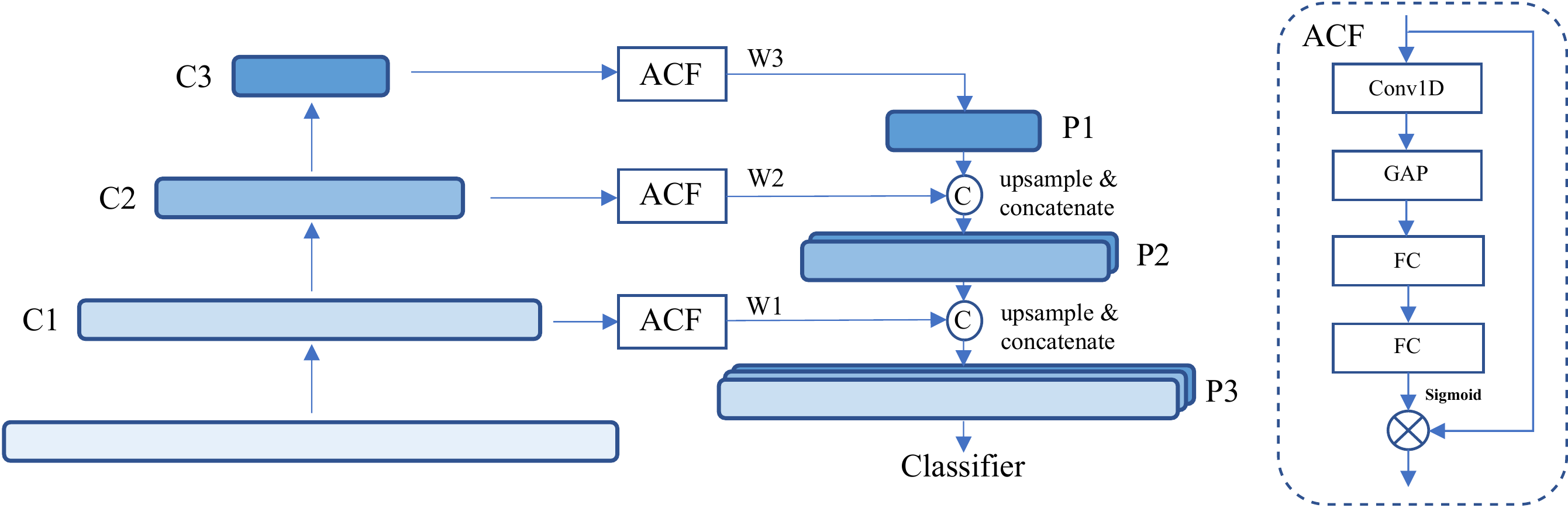}
		\caption{Multi-scale feature fusion model. Specifically, features $\left\lbrace \text{C1,C2,C3} \right\rbrace$   with lengths of 768, 384, 192 are extracted from the $5th$, $7th$, and $9th$ residual blocks of the backbone.
			$\left\lbrace \text{W1,W2,W3} \right\rbrace$ are formed after $\left\lbrace \text{C1,C2,C3} \right\rbrace$ pass through the Adaptive Channel Fusion (ACF) module.
			After upsampling, $\text{P1}$ is concatenated with the shallower feature $\text{W2}$ to form $\text{P2}$; similarly, we get $\text{P3}$, and finally feed $\text{P3}$ into the classifier.} \label{fig2}
	\end{figure}

	In computer vision, Feature Pyramid Network (FPN)  \cite{Lin2017} has shown the promising capability of multi-scale representation by establishing shortcut pathways for shallow features.  This paper designs a multi-scale feature fusion model on the basis of the architecture of FPN to improve our learning ability. The architecture is shown in Fig. \ref{fig2}.
	Different from the original FPN, we adopt channel concatenation to achieve multi-scale feature fusion. 	The main reason for this change is that the channel concatenation preserves more information and avoids the interaction of features with the semantic representation divergence.
	
	Usually, there are common information and unique information among the features of different levels.
	After three-way concatenation, their common information is repeated three times, while the unique information of different levels appears only once. 
	Inspired by the Channel Attention mechanism of SENet  \cite{Hu2020}, we introduce the Adaptive Channel Fusion (ACF) module (as shown in Fig. \ref{fig2})  to weaken the common features and enhance the unique features. In this way, information redundancy is avoided  and  the effectiveness of fusion is improved. 	
	
	The original features with the shape of $L \times C$ are entered into the Adaptive Channel Fusion module, which go through a 1d-convolutional layer, a Global Average Pooling layer, and two fully connection layers of different activation functions (ReLU and Sigmoid).
	In this way, the network learns the weight for each channel, and then the $1 \times C$ weights are multiplied with  the original $L \times C$ input . 
	
	
	Multi-scale feature fusion model fuses features of different depths, which reduces the loss of information and provides a better feature representation for the network as the feature quality is guaranteed.

	\subsection{Markov-based Sequential Correction Algorithm}
\begin{figure}[t]
		\centering
		\subfigure[Ground truth]{\includegraphics[width=0.3\textwidth]{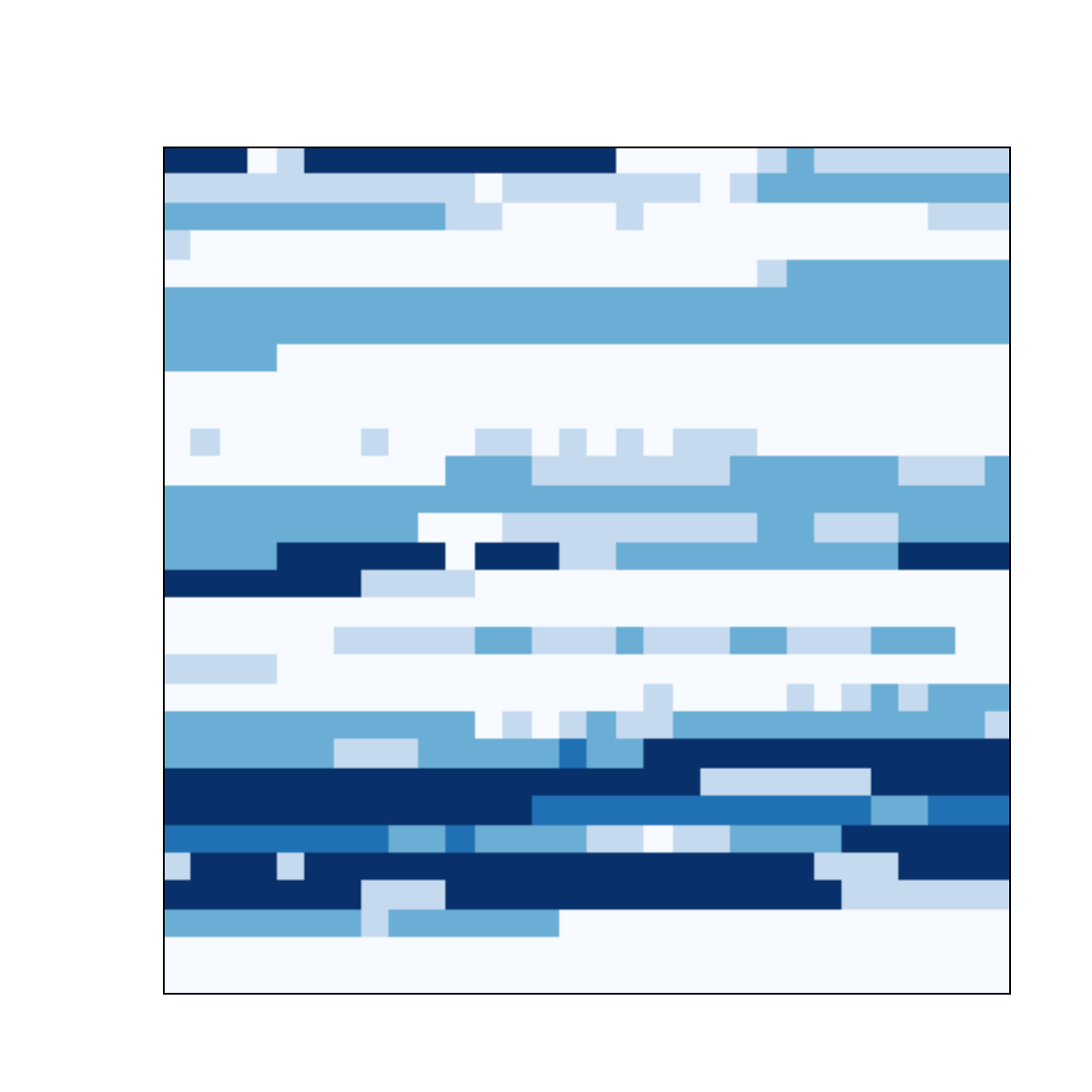}}
		\subfigure[Raw predictions]{\includegraphics[width=0.3\textwidth]{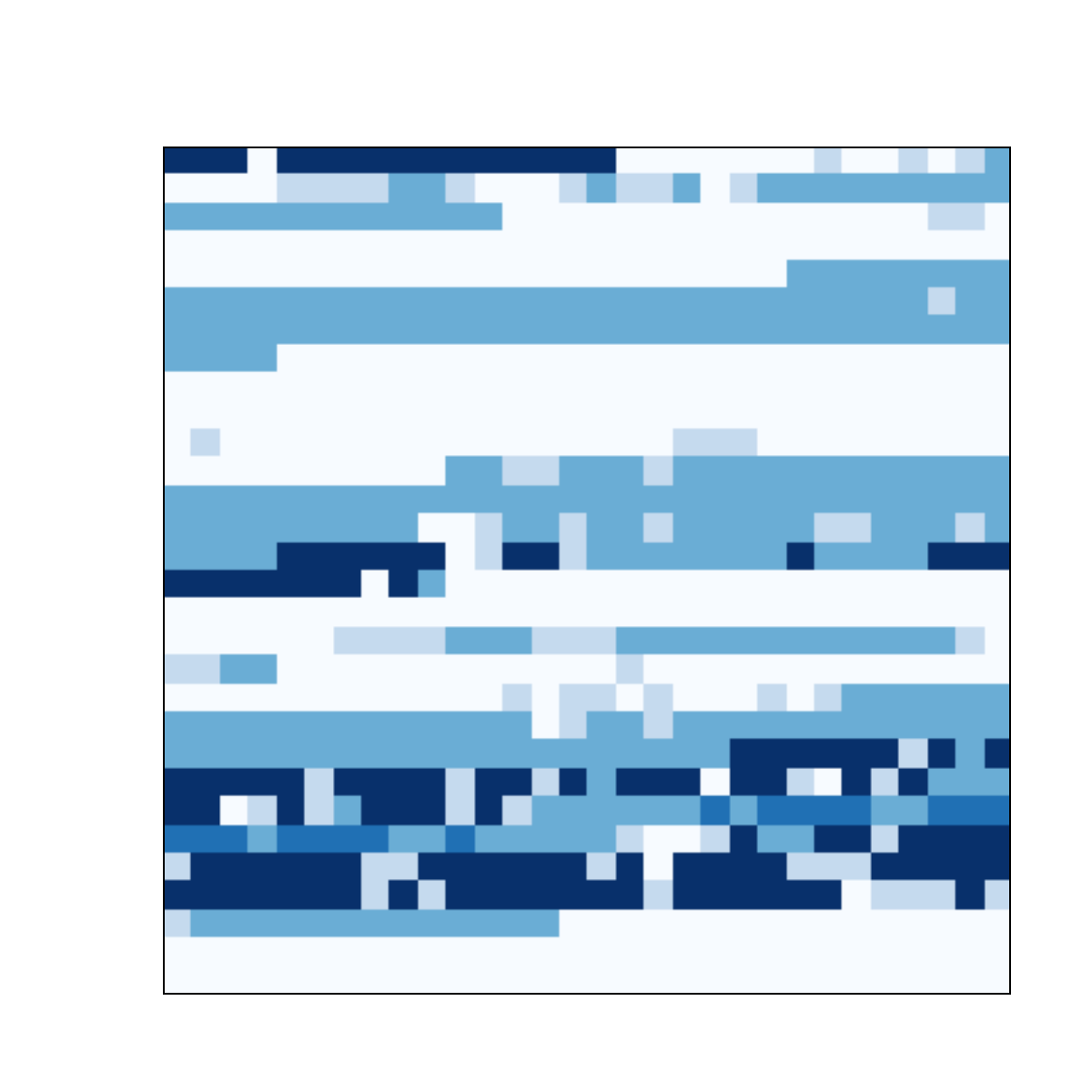}}
		\subfigure[Processed predictions]{\includegraphics[width=0.3\textwidth]{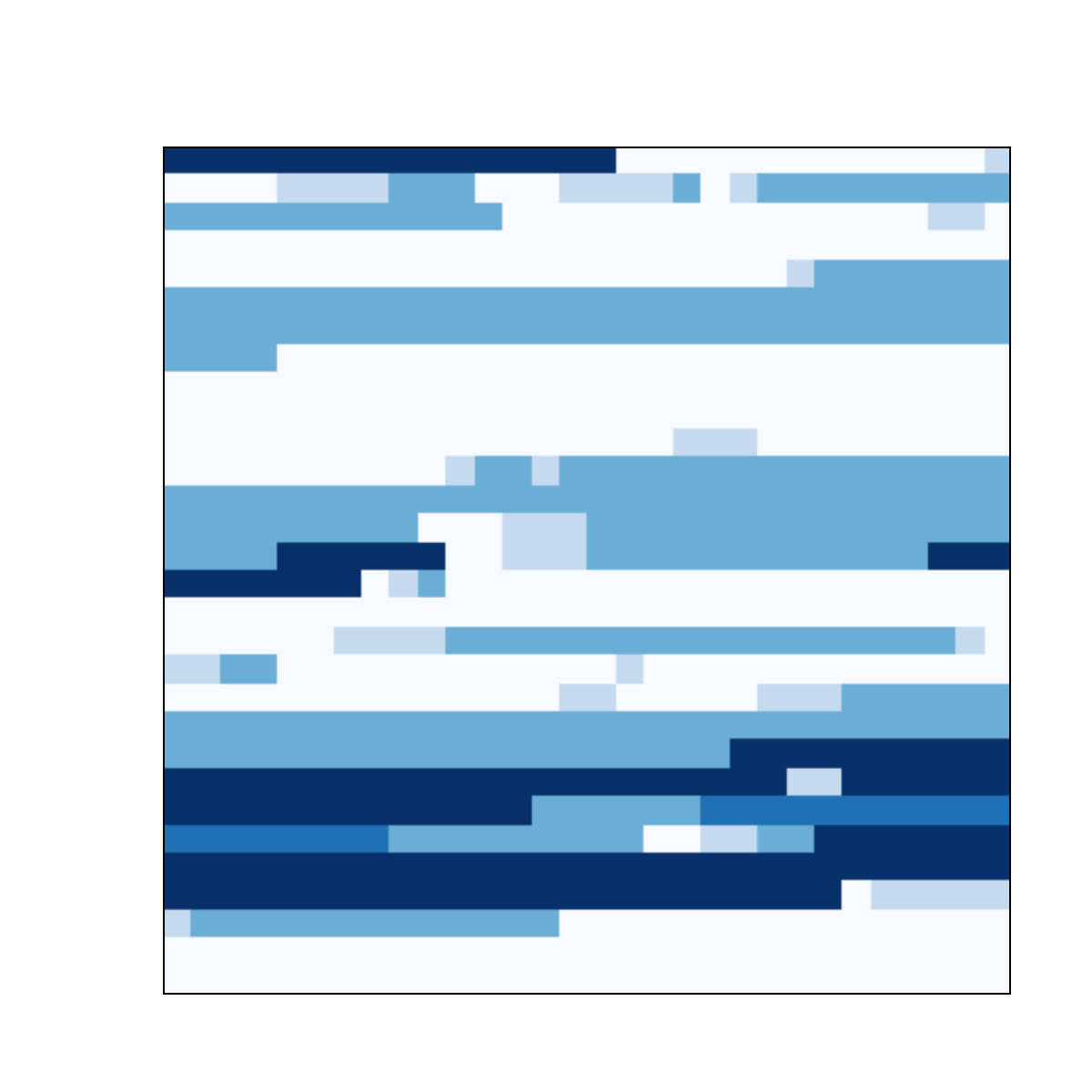}}
		\caption{Comparison of  (a)ground truth, (b)raw predictions, (c)processed predictions of 450-minute signals shows that the output jitters decrease a lot after the Markov-based sequential  correction algorithm. Each grid in the figure represents the stage of each sample, and the five states are distinguished by color shade. } \label{compare}
	\end{figure}

	In the previous sections, the network regards each input signal as an independent and identically distributed sample.
	So the classifier can be regarded as a memoryless source, and the result of sleep scoring is shown in Fig. \ref{compare}(b).
	Compared with the ground truth (Fig. \ref{compare}(a)), the classification results contain a lot of noises. Especially during the continuous sleep stages, output jitters often occurs, i.e., the predictions switch frequently among several stages.
	
\begin{figure}[t]
		\centering
		\subfigure[Transition distribution of train set]{\includegraphics[width=0.45\textwidth]{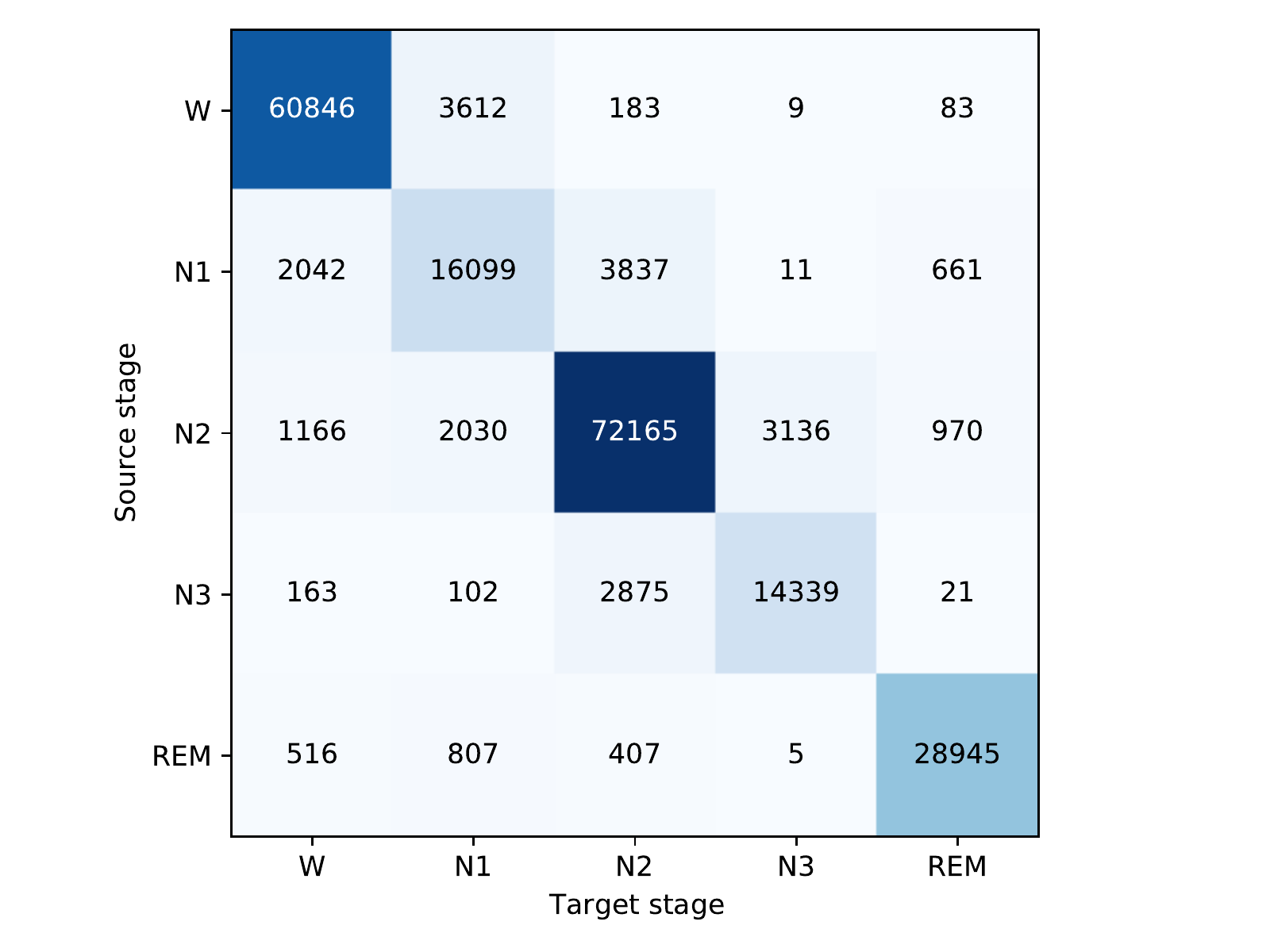}}
		\subfigure[State transition probability matrix ]{\includegraphics[width=0.45\textwidth]{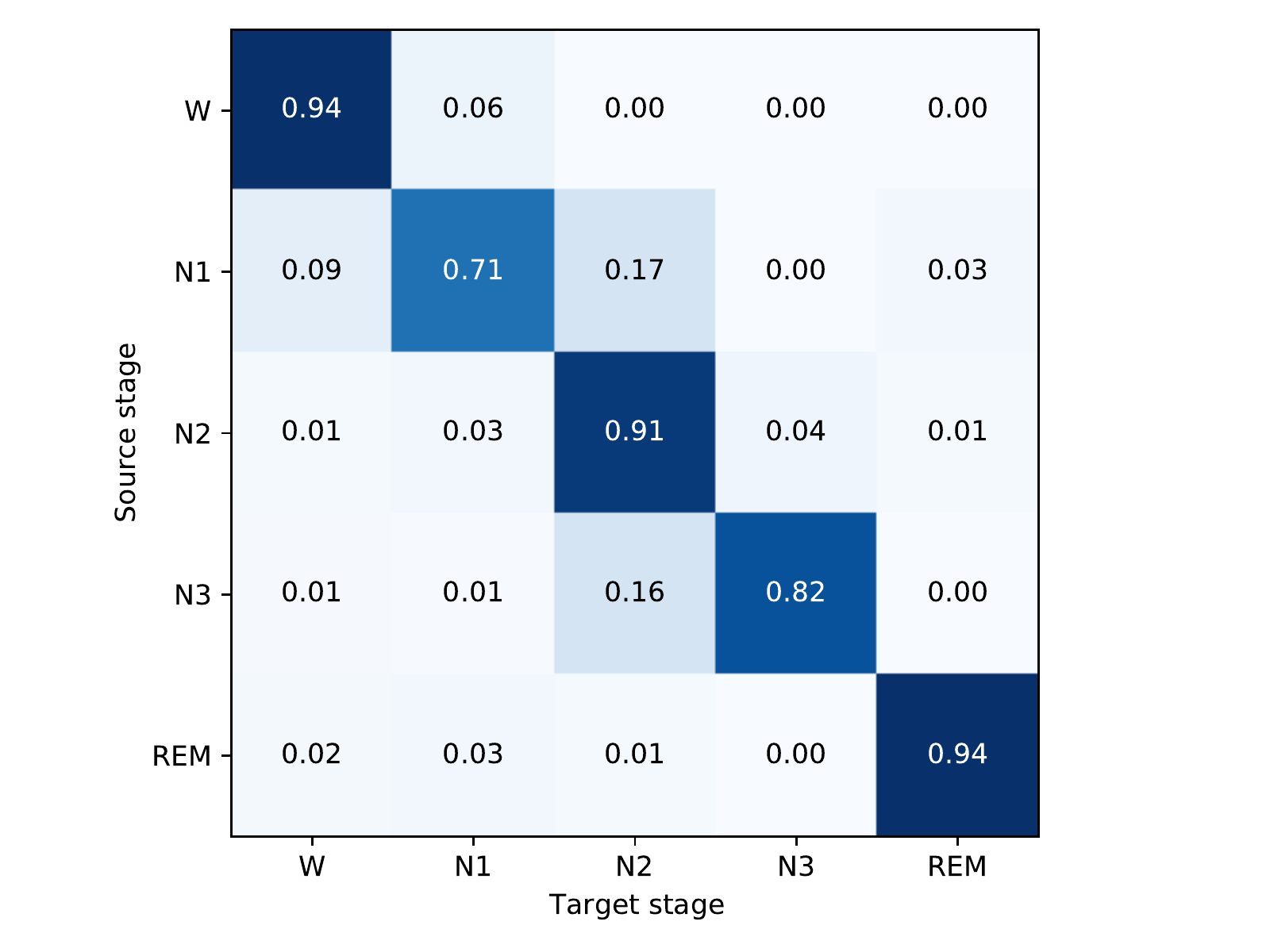}}
		\caption{Two types of transition matrixes of EEG data. The elements represent the (a) number or (b) probability from the source stage (the vertical axis) to the target stage (the horizontal axis).} \label{da}
	\end{figure}

	According to \cite{wiki}, we notice that the change of sleep stages follows a certain rule (called the sleep cycle   \cite{pic}), but many researches  \cite{Perslev2019}\cite{Supratak2017}\cite{Seo2020} ignore this.
	During 7-8 hours of sleep, humans will cycle through these stages in accordance with this rule (e.g., N1$\rightarrow$ N2$\rightarrow $ N3$\rightarrow $ N2$ \rightarrow$ REM) in 90 minutes.
	The transition distribution of EEG data is counted as Fig. \ref{da}(a) and it is obvious that the changes tend to happen between adjacent stages, which follows the rule of the sleep cycle. 
	
	Inspired by sleep cycle, we design a Markov-based sequential  correction (MSC) algorithm for prediction results of the classifier (shown in Algorithm 1 and Fig. \ref{mar}), which can be divided into two parts: 1-order forward checking and $n$-order backward checking. The 1-order forward checking is based on the Markov chain to analyze the relationship between the previous stage and the current stage.
	The $n$-order backward checking further considers the influence of the subsequent $n$ stages on the current stage. 
	
	The Markov-based sequential  correction algorithm is triggered every time the stage prediction changes during a period of sleep. MSC examines the legality of stage changes by weighting the class-wise confidence to filter out those tiny possible stage transitions (e.g., N3 $\rightarrow$ W).
	
	\subsubsection{1-order forward checking}
		\begin{figure}[t]
		\centering
		\begin{minipage}[t]{0.6\textwidth}
			\centering
			\includegraphics[width=0.85\textwidth]{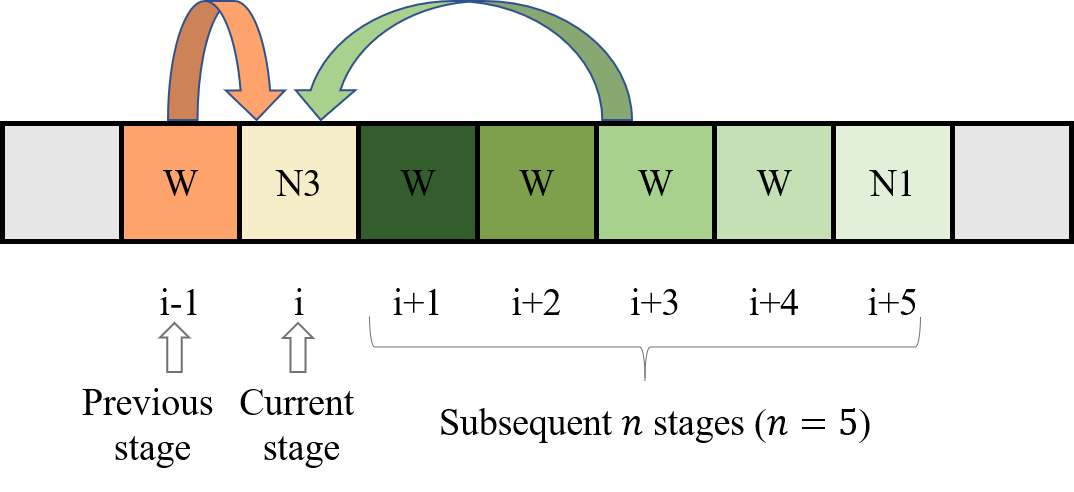}
			\caption{An example of MSC processing. The shade of green represents the degree of influence on the current stage.}\label{mar}
		\end{minipage}
		\hfill
		\begin{minipage}[t]{0.38\textwidth}
			\centering
			\includegraphics[width=\textwidth]{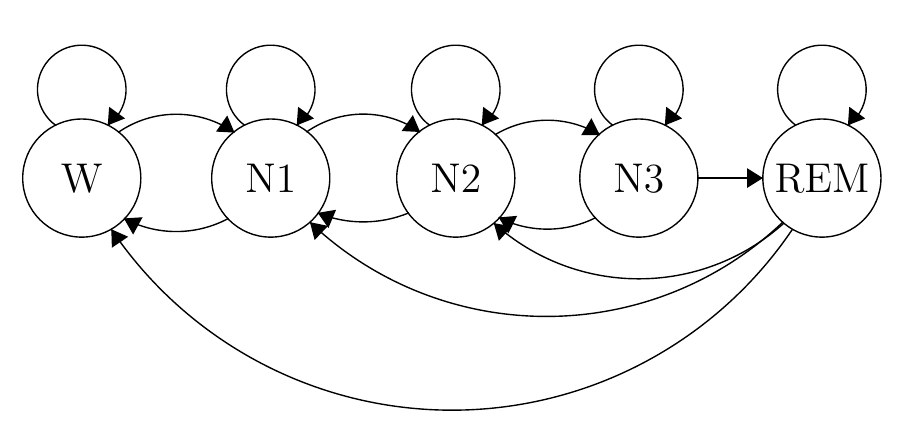}
			\caption{Stage transition graph. It summarizes the relationship between five sleep stages.}\label{stages}
		\end{minipage}
	\end{figure}
	
	In a period of sleep, each segment of EEG signal approximately satisfies the 1-order Markov chain model \cite{Gray2011}, i.e., the probability of each event depends only on the stage attained in the previous event.
	Then the state transition probability matrix (Fig. \ref{da}(b)) is calculated from the transition distribution (Fig. \ref{da}(a)). The vertical axis represents the source stage, the  horizontal axis represents the target stage, and the value is the probability of transition from the source stage to the target stage. The Markov chain in Fig.\ref{stages} summarizes the state  transition probability matrix  (Fig. \ref{da}(b)).
	
	It can be seen from Fig. \ref{da}(b) that the values of the diagonal elements of the matrix are significantly larger than other elements, which shows that the transition of sleep stages rarely occurs during sleep. However, our algorithm acts where the sleep stage changes, so the weight of the diagonal elements should be appropriately compressed to avoid the algorithm from inhibiting the stage transition that was originally predicted to be correct.
	Let $G(\cdot)$ be the compression function, let norm$(\cdot)$ be the normalization by row and the preprocessing for the transition distribution $\bm{M}$  in Fig. \ref{da}(a) is implemented as
	\begin{equation}\label{compress}
	\bm{M'} = \text{norm}(G(\bm{M})^{r}), r \geq 1
	\end{equation}
	where 
$r$ is a constant value and controls the relative size between matrix elements.
	$\bm{M'}$ is the new matrix after preprocessing.
	All the hyperparameters in this section will be specifically discussed  in Section \textbf{3.2}.
	
	
	\begin{algorithm}[t]
		\caption{Markov-based sequential  correction algorithm}
		\begin{algorithmic}[1]
			\Require
			The processd matrix from (\ref{compress}), $\bm{M'}$;		
			
			The $n$-order inertia factor calculated from (\ref{w}), $w_i$;
			
			Classification confidence of the $i$-th sample after softmax layer, $\bm{P_i}$;
			
			The predicted stage index of the $i$-th sample,
			$C_{i}$;
			\Ensure
			MSC processed classification results, ${C'_i}$;
			\For{each sample indexed by i}
			\If {$C_{i} \neq C_{i-1} $}
			\State $\bm{R}$ $\leftarrow $ the  $C_{i-1}$-th row of $\bm{M'}$;
			\State $\bm{R}_{C_{i-1}}$ $\leftarrow$  the  $C_{i-1}$-th element of $\bm{R}$;
			\State $\bm{R}_{C_{i-1}}$ $\leftarrow $ $\bm{R}_{C_{i-1}}$ $\times$ $w_i$;
			\State $\bm{P'_i}$ $\leftarrow $ $\bm{P_i}$ $\times$ $\bm{R}$;
			\State ${C'_i}$ $\leftarrow $ argmax$(\bm{P'_i})$
			\EndIf
			\EndFor	
		\end{algorithmic}
	\end{algorithm}
	
	\subsubsection{$n$-order backward checking}
	The $n$-order backward checking is designed to analyze the impact of the subsequent $n$ stages on the current stage, as shown in Fig. \ref{mar}. 
	Specifically, denoting classification confidence of the $i$-th sample after softmax layer as $\bm{P_i}$, and  the predicted stage index of the $i$-th sample as $C_{i}$, we have 
$	C_{i} = \text{argmax}(\bm{P_i}), C_{i}\in \begin{Bmatrix}{0,1,2,3,4}\end{Bmatrix}.$
	When the output stages of the classifier change, i.e., $C_{i} \neq C_{i-1} $, the algorithm checks subsequent $n$ stages, from $C_{i+1}$ to $C_{i+n}$, to make the classifier more certain about whether the stage has changed or not. This process can be described as
	\begin{equation}\label{w}
	\begin{split}
	w_i =  \sum_{j=1}^{n} \delta_j a^{-j}, (a\geq 1) ,
	\delta_j =\begin{cases} 1& \text{ if } C_{i+j}= C_{i-1} \\ -1 & \text{ if } C_{i+j}= C_{i} \end{cases}, j=1,2,...,n.
	\end{split}
	\end{equation}
	Here, $w_i$ is the inertia factor, representing the degree to which the current stage remains unchanged.  $n$ is the order, representing the number of stages that the algorithm checks after the current stage. $a$ is the decay constant, controlling the weight size of each order. $\delta_j$ represents the change direction of $w_i$.
	The classification confidence is reweighted by $w_i$ as Algorithm 1.

	\section{Experiments}
	\subsection{Dataset and Evaluation Metrics}
	In order to evaluate the sleep scoring performance of our network, we select one of the most commonly used public physiological datasets in the field of automated sleep staging. It has two versions:
	\begin{itemize}
		\item Sleep-EDF  \cite{goldberger2000physiobank}\cite{kemp2000analysis}. It contains PSG records of 20 healthy subjects without sleep-related disorders. The average subject age is 28.7 $\pm $ 2.9 years. Each record provides two-channel EEGs from Fpz-Cz and Pz-Oz channels.
		\item Sleep-EDFx  \cite{goldberger2000physiobank}\cite{kemp2000analysis}. The extend of  Sleep-EDF.  It contains 197 PSG records and consists of two parts: 153 SC records collected from healthy people and 44 ST records collected from people with slight sleep disorders. The number of subjects is 100, whose ages range from 25 to 101.
	\end{itemize}
	
We evaluate the performance of our network using per-class F1 score (F1), overall accuracy (Acc), and macro-averaging F1 score (MF1) \cite{sokolova2009systematic}. 
		The MF1 and Acc are calculated as follows:
	
		\begin{equation}\label{acc}
			\text{Acc} = \frac{1}{N}\sum_{c=1}^{C}\text{TP}_c
		\end{equation}
		\begin{equation}\label{mf1}
			\text{MF1} = \frac{1}{C}\sum_{c=1}^{C}\text{F1}_c
		\end{equation}
		where $\text{TP}_c$ is the true positives of class $c$, $\text{F1}_c$ is per-class F1 score, $C$ is the number of sleep stages and $N$ is the total number of test epochs.
	
	\subsection{Experimental Results}
		\begin{table}[t]
		\centering
		\caption{Performance comparison between MRNet and the state-of-the-art methods for automatic sleep scoring via deep learning. MFF means multi-scale feature fusion model. MSC means Markov-based sequential correction algorithm.  Bold represents the best result under the same conditions.}
		\label{all}
		\resizebox{\textwidth}{!}{%
			\begin{tabular}{cccccccccccccccc}
				\hline
				\multirow{2}{*}{Method} &
				\multirow{2}{*}{Architecture} &
				\multirow{2}{*}{\begin{tabular}[c]{@{}c@{}}Intra-epoch \\ Relation\end{tabular}} &
				\multirow{2}{*}{Dataset} &
				\multirow{2}{*}{Channel} &
				&
				\multirow{2}{*}{Subjects} &
				&
				\multicolumn{2}{c}{Overall Metrics} &
				&
				\multicolumn{5}{c}{Per class F1 score(\%)} \\ \cline{9-10} \cline{12-16}
				&
				&
				&
				&
				&
				&
				&
				&
				Acc(\%) &
				MF1(\%) &
				&
				W &
				N1 &
				N2 &
				N3 &
				REM \\ \cline{1-7} \cline{9-10} \cline{12-16}
				IITNET \cite{Seo2020} &
				CNN-RNN &
				LSTM &
				Sleep-EDF &
				Fpz-Cz &
				&
				20 &
				&
				83.60 &
				76.54 &
				&
				87.10 &
				39.20 &
				87.80 &
				\textbf{87.70} &
				80.90 \\
				DeepSleepNet \cite{Supratak2017}&
				CNN-RNN &
				LSTM &
				Sleep-EDF &
				Fpz-Cz &
				&
				20 &
				&
				82.00 &
				76.88 &
				&
				84.70 &
				\textbf{46.60} &
				85.90 &
				84.80 &
				82.40 \\
				\textit{MRNet} &
				CNN-MFF &
				MSC &
				Sleep-EDF &
				Fpz-Cz &
				&
				20 &
				&
				\textbf{87.59} &
				\textbf{79.62} &
				&
				\textbf{92.35} &
				45.63 &
				\textbf{90.55} &
				86.01 &
				\textbf{83.56} \\ \cline{1-7} \cline{9-10} \cline{12-16}
				DeepSleepNet \cite{Supratak2017} &
				CNN-RNN &
				LSTM &
				Sleep-EDF &
				Pz-Oz &
				&
				20 &
				&
				79.80 &
				73.08 &
				&
				88.10 &
				\textbf{37.00} &
				82.70 &
				\textbf{77.30} &
				80.30 \\
				\textit{MRNet} &
				CNN-MFF &
				MSC &
				Sleep-EDF &
				Pz-Oz &
				&
				20 &
				&
				\textbf{85.10} &
				\textbf{74.11} &
				&
				\textbf{92.65} &
				32.27 &
				\textbf{88.24} &
				74.18 &
				\textbf{83.19} \\ \cline{1-7} \cline{9-10} \cline{12-16}
				U-Time \cite{Perslev2019} &
				Encoder-Decoder &
				None &
				Sleep-EDFx &
				Fpz-Cz &
				&
				78 &
				&
				81.30 &
				76.26 &
				&
				92.03 &
				51.03 &
				83.45 &
				74.56 &
				80.23 \\
				U-Time \cite{Perslev2019} &
				Encoder-Decoder &
				None &
				Sleep-EDFx &
				Fpz-Cz &
				&
				22 &
				&
				83.16 &
				78.61 &
				&
				87.14 &
				\textbf{51.51} &
				86.44 &
				\textbf{84.24} &
				83.70 \\
				\textit{MRNet} &
				CNN &
				None &
				Sleep-EDFx &
				Fpz-Cz &
				&
				100 &
				&
				83.42 &
				77.27 &
				&
				93.54 &
				46.42 &
				85.47 &
				81.11 &
				79.83 \\
				\textit{MRNet} &
				CNN-MFF &
				None &
				Sleep-EDFx &
				Fpz-Cz &
				&
				100 &
				&
				83.63 &
				77.68 &
				&
				93.62 &
				47.47 &
				85.90 &
				81.55 &
				79.84 \\
				\textit{MRNet} &
				CNN-MFF &
				MSC &
				Sleep-EDFx &
				Fpz-Cz &
				&
				100 &
				&
				\textbf{85.14} &
				\textbf{78.91} &
				&
				\textbf{93.71
				} &
				48.07
				&
				\textbf{86.97
				} &
				79.34 &
				\textbf{86.48} \\ \hline
			\end{tabular}%
		}
	\end{table}
	
	The length of the EEG data is 3000 and the input data will be 2$\times$ downsampled many times in the network, so we pad them to 3072. Considering that other networks  \cite{Perslev2019}\cite{Supratak2017}\cite{Seo2020}  adopt cross-validation methods, in order to make our results comparable, 10-fold cross-validation is conducted for both Sleep-EDF and Sleep-EDFx.
	Since the Markov-based sequential  correction algorithm acts on a series of continuous stages, we take continuous  10\% of each record as the test set and the rest as the training set. This is done in order to maintain data continuity.

	We use TensorFlow 2.3 to build MRNet, which is trained on the NVIDIA GTX 1080Ti with the batch size of 128. The network is trained for 70 epochs with random initialization of the weights as described by He et al \cite{he2015delving}.
	We use the SGD optimizer with the momentum $= 0.9$.
	We take 0.1 as the initial learning rate and reduce it by 10 times every 20 epochs.
	
	We compare backbone networks of different depths and different filter lengths. The baseline is the network that performs  best with a depth of 19 and a filter length of 32. The results of the baseline are 83.42\%  Acc and 77.31\%  MF1 on Sleep-EDFx.

	Table  \ref{all} lists the performance of MRNet and the state-of-the-art methods.
	Under the same conditions, MRNet outperforms other methods.
	Experiment results on public dataset Sleep-EDFx and Sleep-EDF show that our proposed approach has achieved high accuracy up to 85.14\%  and 87.59\%, respectively.
	Comparing data from different channels on the same model reveals that the data from the Fpz-Cz channel have a better performance than the  Pz-Oz channel. This may be due to the biased distribution of the two data.
	In addition, it can be found in Table  \ref{all}  that the performance of Sleep-EDFx is worse than that of Sleep-EDF. The main reason for this is that Sleep-EDFx contains more complex data distribution, which makes it more difficult to classify.
	
	MRNet for different structures is also compared in Table  \ref{all}. The multi-scale feature fusion model (MFF) improves overall performance by improving the classification results for each category.
	In contrast, the Markov-based sequential  correction algorithm sacrifices the performance of N3 in exchange for a substantial increase in REM results.
	
	\begin{table}[!t]
		\begin{minipage}[t]{0.5\textwidth}
			\centering
			\caption{Architecture study of multi-scale feature fusion model on Sleep-EDFx. ACF: the Adaptive Channel Fusion module. }
			\label{table2}
			\begin{tabular}{ccccc}
				\hline
				\begin{tabular}[c]{@{}c@{}}Backbone\\ Depth\end{tabular} & \begin{tabular}[c]{@{}c@{}}Feature\\ Pyramid\end{tabular} & ACF & Acc(\%) & MF1(\%) \\ \hline
				19 & 2 ways & w/o & 83.29          & 77.29          \\
				19 & 2 ways & w/   & 83.42          & 77.58          \\
				19 & 3 ways & w/   & \textbf{83.63} & \textbf{77.68} \\
				23 & 2 ways & w/   & 83.52          & 77.05          \\
				23 & 3 ways & w/  & 83.61          & 77.41          \\
				23 & 4 ways & w/   & 83.32          & 76.77          \\ \hline
			\end{tabular}
		\end{minipage}
		\hfill
		\begin{minipage}[t]{0.45\textwidth}
			\centering
			\caption{Parameter comparison on the Markov-based sequential correction algorithm on Sleep-EDFx.}
			\label{table3}
			\begin{tabular}{cccc}
				\hline
				\begin{tabular}[c]{@{}c@{}}Compression \\ Function\end{tabular} & \begin{tabular}[c]{@{}c@{}}Order \\ $n$\end{tabular} & Acc(\%)         & MF1(\%)            \\ \hline
				linear                                                          & 1                                                    & 83.96          & 76.43          \\
				sqrt                                                            & 1                                                    & 85.04          & 78.59          \\
				log                                                             & 1                                                    & 85.01          & 78.87          \\
				log                                                             & 2                                                    & 85.12          & \textbf{78.98} \\
				log                                                             & 4                                                    & \textbf{85.14} & 78.91          \\
				log                                                             & 6                                                    & 85.13          & 78.85          \\ \hline
			\end{tabular}
		\end{minipage}
		
	\end{table}

	\subsubsection{Architecture Study on Multi-scale Feature Fusion Model}
	In Sleep-EDFx, exploiting a suitable architecture of the multi-scale feature fusion model for the representation learning is a key factor for the improvement of the sleep staging performance.
	Table \ref{table2} compares the performance with the backbone depths, the number of feature pyramid ways and the existence of the  Adaptive Channel Fusion module.
	The best architecture is based on the 19-layer backbone and has a 3-way feature pyramid with an Adaptive Channel Fusion module, whose results are 83.63\% Acc and 77.68\% MF1.
	From the results, it is obvious that the performance of the model with the  Adaptive Channel Fusion module is $+0.16\%$ Acc and $+0.29\%$ MF1 than a model without the module.


	\subsubsection{Parameter Selection on Markov-based Sequential  Correction Algorithm }
	For the Markov-based sequential  correction algorithm, we mainly focus on the selection of the compression function $G(\cdot)$ in  (\ref{compress}) and the order $n$ for the inertia factor $w_i$ in  (\ref{w}).
	The value of $r$ in  (\ref{compress}) is chosen as 4.2 through grid search.
	In the same way, $a$ in  (\ref{w}) is chosen as 1.5.
	It can be observed in Table  \ref{table3} that the log function is the best choice compared to the other function under the same conditions.
	When the order $n$ gradually increases, the result shows a trend of first increasing and then decreasing. The best accuracy is 85.14\% at $n=4$ and the highest MF1 is 78.98\% at $n=2$. Finally, we choose $n=4$ for the order hyperparameters.

	\section{Conclusion}
	We propose MRNet to reduce the information loss for sleep staging. Specifically, we first adopt the residual block-based network as the backbone to recover the information upper boundary of output features. Then, we design a multi-scale feature fusion model to fuse deep and shallow features from the backbone, thus making the network look at the signals in different scales.
	In particular, we notice that the output jitters seriously interferes with the performance of the classifier. Hence, we propose the Markov-based sequential correction algorithm to take the time series relation of EEG into consideration, which smooths the prediction results.  Compared with the state-of-the-art models, MRNet performs better on both Sleep-EDF and Sleep-EDFx dataset.

	\bibliographystyle{splncs04}
	\bibliography{MyCollection}
\end{document}